\documentclass[sigconf,9pt,authorversion]{acmart}

\usepackage{amsmath,amsfonts}
\usepackage{algorithmic}
\usepackage{graphicx}
\usepackage{textcomp}
\usepackage{xcolor}
 
\usepackage{booktabs}
\usepackage{tabularx}
\usepackage[linesnumbered,ruled,vlined]{algorithm2e}
\usepackage{adjustbox} 
 \usepackage{enumitem}
\usepackage{booktabs} 
\usepackage{makecell} 

\usepackage{todonotes}
\def\BibTeX{{\rm B\kern-.05em{\sc i\kern-.025em b}\kern-.08em
    T\kern-.1667em\lower.7ex\hbox{E}\kern-.125emX}}
\newcommand{\linelabel}[1]{\label{#1}}

\copyrightyear{2024} 
\acmYear{2024} 
\setcopyright{rightsretained} 
\acmConference[EdgeSys '24]{7th International Workshop on Edge Systems, Analytics and Networking}{April 22, 2024}{Athens, Greece}
\acmBooktitle{7th International Workshop on Edge Systems, Analytics and Networking (EdgeSys '24), April 22, 2024, Athens, Greece}

\renewcommand\footnotetextcopyrightpermission[1]{} %
 
\begin{document}

\title{ FLIGAN: Enhancing Federated Learning with Incomplete Data using GAN}  

\author{Paul Joe Maliakel}
\affiliation{%
  \institution{Vienna University of Technology}
  \country{}
  }
\email{paul.maliakel@tuwien.ac.at}

\author{Shashikant Ilager}
\affiliation{%
  \institution{Vienna University of Technology}
  \country{}
  }
\email{shashikant.ilager@tuwien.ac.at}

\author{Ivona Brandic}
\affiliation{%
  \institution{Vienna University of Technology}
  \country{}
  }
\email{ivona.brandic@tuwien.ac.at}

\begin{abstract}
Federated Learning (FL)  provides a privacy-preserving mechanism for distributed training of machine learning models on networked devices (e.g., mobile devices, IoT edge nodes). It enables  Artificial Intelligence (AI) at the edge by creating models without sharing actual data across the network. 
Existing research typically focuses on generic aspects of non-IID data and heterogeneity in client's system characteristics, but they often neglect the issue of insufficient data for model development, which can arise from uneven class label distribution and highly variable data volumes across edge nodes. In this work, we propose FLIGAN, a novel approach to address the issue of data incompleteness in FL.  First, we leverage Generative Adversarial Networks (GANs) to adeptly capture complex data distributions and generate synthetic data that closely resemble real-world data. Then, we use synthetic data to enhance the robustness and completeness of datasets across nodes. Our methodology adheres to FL's privacy requirements by generating synthetic data in a federated manner without sharing the actual data in the process. We incorporate techniques such as classwise sampling and node grouping, designed to improve the federated GAN's performance, enabling the creation of high-quality synthetic datasets and facilitating efficient FL training. Empirical results from our experiments demonstrate that FLIGAN significantly improves model accuracy, especially in scenarios with high class imbalances, achieving up to a 20\% increase in model accuracy over traditional FL baselines.

\end{abstract}

\maketitle
 
\section{Introduction}

The growth of edge computing is propelled by the demand for latency-sensitive smart applications across various domains such as healthcare, smart cities, connected autonomous vehicles, and smart digital assistants ~\cite{oueida2018edge,  liu2019intelligent, liu2019edge}, among others.  Such latency-sensitive smart applications vastly depend on Machine Learning (ML) or Deep Learning (DL) models that provide real-time inference based on input data. However, due to the need for data privacy and limited network resources, developing ML models in a centralized cloud has become infeasible ~\cite{chen2012data} requiring a new paradigm in model development known as  Federated Learning (FL) \cite{yang_2019_federated}. FL is a decentralized ML paradigm that facilitates collaborative model training without necessitating the sharing of raw data ~\cite{li2020federated}. Multiple edge nodes (federated clients) can actively participate in the training process, contributing their local data, all while preserving the confidentiality of sensitive information. This collaborative approach holds the promise of developing sustainable and accurate models that preserve data privacy and network bandwidth.

One of the main challenges of FL is data incompleteness ~\cite{zhao2018federated}, which can be defined with two scenarios: (i) Certain classes predominate on some nodes while absent on others, and/or (ii) Data volume varies across nodes, with some possessing more data than others. The problem of data incompleteness is exacerbated in edge environments, as edge nodes often operate in unreliable conditions and can experience various issues such as temporal device failures, sensor malfunctioning, and network connectivity issues ~\cite{mcmahan2017communication,yang_2019_federated}.

Incomplete data poses significant challenges for developing accurate FL models ~\cite{zhao2018federated}. Incomplete data can affect model robustness, as the aggregated updates from nodes may produce models that are overfitted to their specific test cases,  reducing generalization to unseen data. Moreover, it can lead to biased models, especially when some class labels are underrepresented. Such models can have serious consequences in application domains where fairness and unbiased predictions are crucial \cite{brisimi2018federated}. Common strategies to address incomplete data, such as \textit{imputation} or \textit{oversampling} of missing or scarce class labels \cite{jo2004class, chawla2002smote}, may not be applicable for nodes with sparse data. Furthermore, these methods generally work well in continuous time series data and can cause errors in tabular data, affecting the model's quality and accuracy. Most importantly, methods like \textit{imputation} across nodes require data sharing, which violates the privacy requirements of FL. Thus, in this work, we focus on enhancing the FL model with privacy preservation using incomplete tabular datasets.

Several studies have addressed different issues related to FL, such as node heterogeneity \cite{li2020federated}, model drifting and staleness \cite{bonawitz2019federated}, and system parameters such as node reliability, energy and bandwidth \cite{lim2020federated}. However, they do not explicitly address the problem of data incompleteness, especially for tabular data in FL settings. To overcome this challenge, we propose the use of Generative Adversarial Networks (GANs), which provide a sophisticated approach to generate synthetic data that can effectively augment incomplete datasets while preserving user privacy. GANs consist of two main components: a generator and a discriminator. The generator's role is to produce synthetic data, while the discriminator's role is to evaluate this data, distinguishing between generated and real samples. Unlike simpler \textit{oversampling} and \textit{imputation} techniques, GANs can capture complex data distributions, offering an efficient solution without compromising data integrity.

While GANs have traditionally been applied in the area of image and text synthesis \cite{chang2020synthetic, mcmahan2017learning}, \textbf{our work explores their capability to generate tabular data within an FL framework in a privacy-preserving manner.} Directly applying existing federated GAN models for decentralized tabular data presents some challenges. A significant issue is the multimodal distributions in continuous columns across nodes, leading to mode collapse and widening the discrepancy between synthetic and real data \cite{xu2019modeling}. Thus, in this work, we propose  FLIGAN (\textbf{F}ederated \textbf{L}earning with \textbf{I}ncomplete data using \textbf{GAN}), which utilizes GANs in federated settings and addresses these challenges by adopting several techniques including classwise sampling and node grouping by clustering nodes based on their data quantity. This technique trains generators separately for each class label, thereby narrowing the gap between synthetic and real data in the process, as observed from the results. Furthermore, we use these generators in runtime and add synthetic data step by step in the training process of the FL classification model.  Our experimental study on real-world datasets demonstrates that FLIGAN outperforms baselines by improving model accuracy.

Section \ref{sec:background} outlines the study's background, \ref{sec:ourstudy} introduces our FLIGAN framework, \ref{ref:experiments} details experiments and results analysis, \ref{sec:relatedwork} reviews related work, and  \ref{sec:conclusion} presents conclusions.

\section{Background}\label{sec:background}
\subsection{Federated learning and data incompleteness}

\textbf{Federated Learning (FL)} works on a decentralized model where edge devices collaborate with a central server to train a global model without sharing local data—only model updates are exchanged. The objective is to minimize the global function $F(w)$, representing the weighted aggregation of local functions across devices:

\begin{equation}
    \min_w F(w) = \sum_{k=1}^m p_k F_k(w)
\end{equation}

where $F(w)$ is the global objective, $k$ is the total number of clients participating in the FL process, $w$ is model parameters, $F_k(w)$ is the local objectives on $m$ devices, and $p_k$ their respective weights.

\textbf{Data incompleteness} in FL originates from the problem of missing or insufficient data for some classes in the distributed data sets of the edge nodes.  
Thus, the concept of \textit{data incompleteness} has two primary dimensions,  \textit{ (i) Uneven Class Distributions:} where data classes across nodes are highly variant, for instance, some classes are predominant on some nodes and absent on others; \textit{ (ii) Uneven Data Volume Distributions:} the amount of data varies across nodes where some have more data than others. Such occurrences reflect unique user behaviors and operational conditions leading to data incompleteness. Both aspects of the incompleteness of the data present distinct challenges during model training, and it can lead to poor or biased model performance.

\par 
It is important to note that we focus on data incompleteness, different from the generic case of a non-IID problem which is characterized by underlying data distribution \cite{zhao2018federated}. For instance, the same class of images from two devices could have variations in lighting or angles, leading to unique data distributions. Thus, we assume existing FL algorithms deal with generic aspects of non-IID and we focus on addressing the lack of insufficient data in FL.

\subsection{GANs}
The Generative Adversarial Network (GAN) is an ML framework that trains to generate realistic data \cite{goodfellow2014generative}. GANs consist of two neural networks, the \emph{generator} and the \emph{discriminator}. The generator aims to create synthetic data resembling real data, while the discriminator aims to differentiate between real and generated samples. In an iterative process, the generator improves on producing convincing data, while the discriminator becomes more adept at distinguishing real from fake data. Despite their capability, GAN training can be challenging due to issues like mode collapse and instability \cite{xu2019modeling}. 


While numerous GAN variants tackle limitations of traditional GANs, Wasserstein GAN (WGAN) \cite{arjovsky2017wasserstein} addresses these 
by using Wasserstein distance as a more reliable dissimilarity measure between real and generated distributions. WGAN-GP \cite{gulrajani2017improved} improves on WGAN by adding a gradient penalty to maintain Lipshitz continuity more effectively than weight clipping, stabilizing training, and generating higher-quality samples. Therefore, we choose WGAN-GP architecture for our problem. In the next section, we present our FLIGAN system model and discuss the proposed algorithms.

\section{FLIGAN: Federated Learning with Incomplete Data using GAN}\label{sec:ourstudy}
\begin{figure*}[!ht]
    \centering
    \includegraphics[width=1\textwidth]{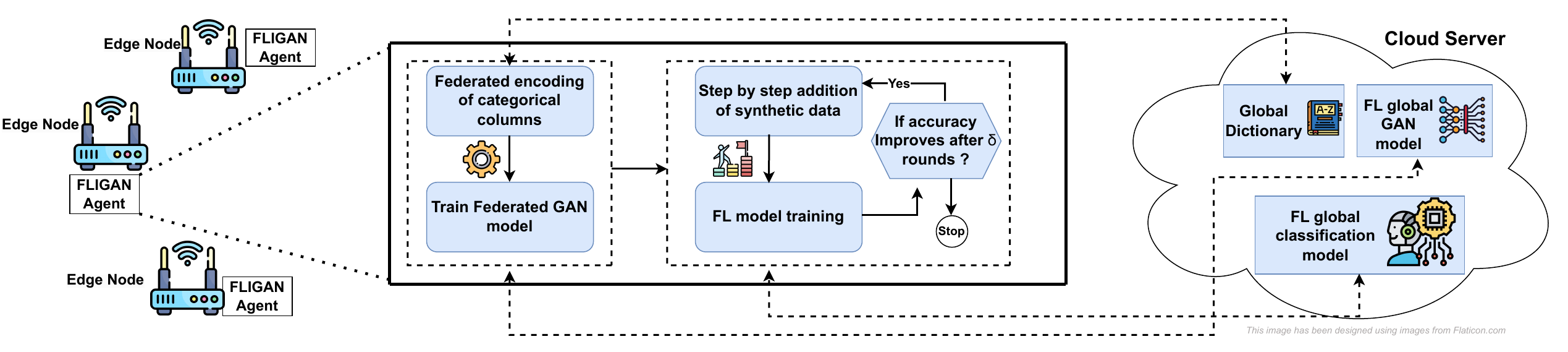}
    \caption{A high-level overview of the FLIGAN system model}
    \label{fig:methodology}
    \end{figure*}

Figure \ref{fig:methodology} shows our system model explaining the workflow FLIGAN.  \textit{First}, we develop a federated GAN model that is trained without sharing the edge node's data. Second, we use the trained GAN model to generate new synthetic data for each edge node. Then, we incrementally add the synthetic data to train a federated model, which results in an enhanced FL model compared to a model trained with incomplete data. In the following subsections, we explain the components of this system model in detail.

\subsection{Developing federated GAN model for synthetic data generation}

In this section, we first describe federated encoding for categorical columns, as it is essential for all nodes to have a consistent representation of categorical data. Then, we discuss how we develop the federated GAN model and, ultimately, the FL classification model. 

\subsubsection{Federated encoding of categorical columns}
Handling incomplete tabular data with categorical columns is challenging as some nodes may not have a representation of certain categorical values. However,  without global knowledge of the unique categorical values of columns, each node might encode categorical columns differently, creating inconsistencies in model aggregation and training. 
To counter this, we employ a decentralized federated encoding strategy to ensure uniform data representation, creating a shared global dictionary of categorical values of respective columns from all nodes' data. This dictionary is then shared among nodes. The process starts with the nodes submitting their categories metadata (unique values) to a central server, which then compiles a global dictionary for distribution, allowing for uniform encoding of categorical variables system-wide and ensuring a consistent data representation across the network.

\subsubsection{Training federated GAN model with classwise sampling \& node grouping}

The Algorithm \ref{alg:fedgan} of our federated GAN model starts with nodes sharing their data's class distribution with the server (line \ref{alg:line:class}). The server then initializes the global generator ($G_{label}$) and discriminator ($D_{label}$) models to unify the learning base across the nodes (line \ref{alg:line:ini}) using the Wasserstein GAN with Gradient Penalty (WGAN-GP) for effective training.

The training phase of our federated GAN model involves iteratively grouping nodes by the data volume for each class label, using the DBSCAN algorithm  \cite{ester1996density} to cluster based on data density (line \ref{alg:line:group}). This results in node groups with similar data amounts, prioritized by data volume for targeted training. To prevent overfitting and optimize resources, training rounds, and epochs are adjusted exponentially for each group, guided by decay rates $\alpha_R$ and $\alpha_E$ (line \ref{alg:line:rounds} \& \ref{alg:line:epochs}). This approach ensures efficient, balanced training across diverse node datasets.

In our federated GAN model's training, we start with the most data-rich edge node group for each class label and train for initial \textit{Rounds} and \textit{Epochs} to maximize data representation. Training \textit{Rounds} and \textit{Epochs} then decrease exponentially for groups with less data, preventing overfitting on sparse datasets. Nodes train local GAN models on their specific data, aggregating the model weights with the server after each round. This process iterates across all class labels for each of the particular label's node groups, and  final models are distributed back to all nodes, completing the FL cycle (line \ref{alg:line:dist}). The result is a set of generator models for each class label, capable of generating synthetic data, helping to mitigate the class imbalance issues in training iterations.  Our algorithm uniquely addresses data imbalances and overfitting risks with classwise sampling, making it versatile for diverse federated settings.

\begin{algorithm}[!t]
\caption{ Federated GAN for synthetic data generation}
\label{alg:fedgan}
\DontPrintSemicolon
\KwData{Initial Rounds \( R_{init} \), Initial Epochs \( E_{init} \), Class Labels \( L \), Decay Rates \( \alpha_R \), \( \alpha_E \)}
\KwResult{EncodedData, Generators \( \{G_{label}\} \)}

\BlankLine
EncodedData \( \leftarrow \) FederateEncode(IncompleteData)\;
\ForEach{Node}{
    Send class distribution of local data to central server\;\linelabel{alg:line:class}
}
\For{\( label \) in \( L \)}{ 

    Initialize \( G_{label} \), \( D_{label} \)\; \linelabel{alg:line:ini}
    NodeGroups \( \leftarrow \) SortDescending(GroupNodes(\( label \)))\; \linelabel{alg:line:group}
    
    \For{groupIdx \( \leftarrow \) 0 \KwTo \( |NodeGroups| - 1 \)}{
        Rounds \( \leftarrow \) \( \lceil R_{init} \cdot \alpha_R^{\text{groupIdx}} \rceil \)\; \linelabel{alg:line:rounds}
        Epochs \( \leftarrow \) \( \lceil E_{init} \cdot \alpha_E^{\text{groupIdx}} \rceil \)\; \linelabel{alg:line:epochs}
        \For{\( r = 1 \) \KwTo Rounds}{  
            \For{client \( c \) in NodeGroups[groupIdx]}{
                Train WGAN-GP(\( G_{local_c} \), \( D_{local_c} \), LocalData\(_{c,label} \), Epochs)\;
                Send weights to server \; \linelabel{alg:line:train}
            }
            Update \( G_{label} \), \( D_{label} \)\;
        }
        
    }
    Distribute \( G_{label} \)\; \linelabel{alg:line:dist}
}
\end{algorithm}

\subsection {Training FL classification model with GAN generated data}

In this section, we explain how we use the federated GAN model developed in the previous section to train an actual FL classification model. The pseudocode for training the federated classification model using the data generator is shown in Algorithm \ref{alg:fligan}.

\begin{algorithm}[t]
\caption{FLIGAN}
\label{alg:fligan}
\DontPrintSemicolon
\SetKwInOut{Input}{Input}
\SetKwInOut{Output}{Output}
\SetKwInOut{Initialize}{Initialize}
\SetKwFor{For}{for}{do}{end}
\SetKwIF{If}{ElseIf}{Else}{if}{then}{else if}{else}{endif}
\SetKw{Return}{return}

\Input{IncompleteData}
\Output{FL classification model}
\BlankLine
\Initialize{
    EncodedData, Generator $\leftarrow$ train\_data\_generator(IncompleteData)\;
    CombinedData $\leftarrow$ EncodedData 
}
\Repeat{AccuracyNotImprovedAfter{\_$\delta$\_rounds}}{ 
    SyntheticData $\leftarrow$ StepByStepAddition(Generator)\;  \linelabel{alg:line:step}
    CombinedData $\leftarrow$ Merge(CombinedData, SyntheticData)\;  
    Model $\leftarrow$ TrainFederatedModel(CombinedData)\;  \linelabel{alg:line:model}
}\linelabel{alg:line:if}
\end{algorithm}

\textbf{Step-by-step addition of synthetic data}:
This method involves adding synthetic data to node datasets in a step-by-step manner to balance class representation (line \ref{alg:line:step}). We begin by identifying the class with the maximum number of samples across nodes, and then a small percentage of this maximum value of synthetic data is added to less represented classes in iterative steps. After each addition, the FL model is trained to evaluate performance improvements (line \ref{alg:line:model}). It continues iteratively until no further gain is observed for $\delta$ (stopping criterion) previous rounds  (line \ref{alg:line:if}). By incrementally enhancing datasets, we aim for a balanced class distribution without overfitting, therefore optimizing the model's accuracy with an ideal amount of synthetic data.

\textbf{FL model training}: Upon adding a percentage of synthetic data to each node, we train the federated classification model using the FedAvg algorithm \cite{mcmahan2017communication}, aiming to a model that generalizes well across diverse data distributions. This approach effectively tackles data incompleteness, showcasing the adaptability of FL to ensure robustness under such conditions of incompleteness.

\section{Performance Evaluation}\label{ref:experiments}

In this section, we provide the details of implementation, experimental design, and analysis of the results. 

\subsection{Experimental setup} 
\textbf{Implementation.} We implemented FLIGAN using the Flower framework 1.4.0 \cite{flowerFramework},  a generic framework to develop FL systems. We use PyTorch 1.13.1. to implement GAN and classification DL models.  We emulate our experimental settings on a single node, including edge nodes and a cloud server.
\\
\textbf{Datasets. } Our methodology's effectiveness is assessed using five notable tabular datasets—Adult \cite{Dua:2019}, Intrusion \cite{Tavallaee2009}, Creditcard \cite{misc_default_of_credit_card_clients_350}, Bank (specifically a subset to address class imbalance) \cite{misc_Bank_marketing_222}, and Albert \cite{openml_Albert_2023} as outlined in Table \ref{tab:datasets} with their primary attributes.
\\
\textbf{Generating incomplete data settings.} We use Dirichlet distribution \cite{ng2011dirichlet} to simulate realistic settings for  data incompleteness. Generally, Dirichlet distribution is used to create different types of non-IID conditions, here we adjust Dirichlet distribution parameters to simulate various patterns of data incompleteness across edge nodes. By varying the Dirichlet concentration parameters, we enable a systematic exploration of FL performance under various level of data incompleteness.

\begin{table}[!t]
\centering
\begin{tabular}{|l|c|c|c|c|}
\hline
\textbf{Data} & \textbf{Rows} & \textbf{Cat} & \textbf{Cont} & \textbf{Tot} \\
\hline
Adult & 49k & 9 & 6 & 15 \\
Intrusion & 25k & 4 & 38 & 42 \\
Creditcard & 13k & 8 & 14 & 22 \\
Bank & 10k & 11 & 6 & 17 \\
Albert & 58k & 9 & 23 & 32 \\
\hline
\end{tabular}
\caption{Summary of datasets}
\label{tab:datasets}
\end{table}

Our experiments vary Dirichlet alpha (0.05, 1.0, 1.5, 2.0) across 8 nodes to generate diverse incomplete data scenarios, from severe class imbalances to more uniform distributions. Lower alpha values represent extreme data sparsity, while higher values indicate less incompleteness. Each scenario was repeated thrice for consistency, with analyses based on average accuracy for comparative plots.
\\
\textbf{Baselines.}
We use the following two baselines against FLIGAN.
\begin{itemize}[leftmargin=*, noitemsep,topsep=0pt]
\item \textbf{FedAvg:} A foundational FL algorithm  as a standard benchmark.

\item \textbf{FedGAN:} Unlike FLIGAN, which trains separate GAN models per class label and groups nodes by sample count similarity, FedGAN utilizes the entire dataset of each node for federated GAN training. It shares the WGAN-GP method and neural architecture with FLIGAN, allowing direct performance comparisons. In our study, we substitute FLIGAN's GAN with FedGAN to assess and compare the outcomes.

\end{itemize}
For all experiments, the FL classification model is trained for 10 rounds. The GAN in FedGAN undergoes training for 5 rounds and 60 epochs. FLIGAN initiates with 3 rounds (\(R_{\text{{init}}}\)) and 60 epochs (\(E_{\text{{init}}}\)), with a stopping criterion (\(\delta\)) set to 2.

\subsection{Results and analysis}

\subsubsection{\textbf{Analysis of  the  FLIGAN performance}}

\begin{figure}[!t]
    \centering
    \includegraphics[scale=0.57]{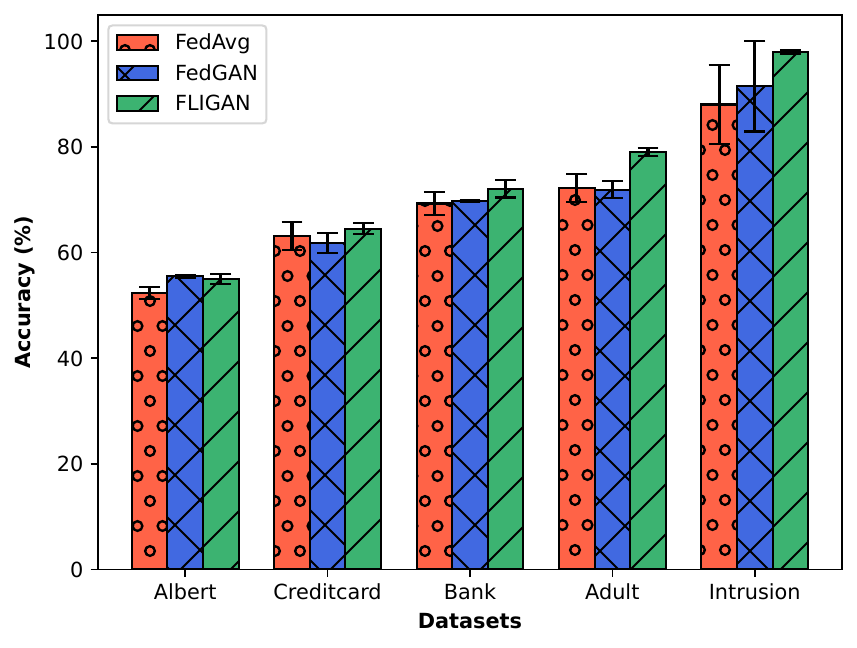}
    \caption{Performance of different algorithms: Averaged accuracy of the FL classification model across different datasets using different techniques.}
    \label{fig:acc}
\end{figure}

When it comes to model accuracy as depicted in Figure \ref{fig:acc}, FLIGAN stands out, particularly in the \textit{Intrusion} dataset where it achieves the highest accuracy (up to 20\% increase) among the strategies. This indicates that FLIGAN, despite its computational demands, is capable of producing highly effective models, especially in cases where dataset characteristics are similar to those in \textit{Intrusion} dataset. The performance on the \textit{Albert} dataset is less impressive, which could be due to the nature of the data or the suitability of the strategy to the dataset.

 \begin{table}[!t]
\centering
\begin{tabular}{|c|c|c|c|}
\hline
\textbf{Dataset} & \textbf{FedAvg} & \textbf{FedGAN} & \textbf{FLIGAN} \\
\hline
Albert & 87 & 1787 & 1754 \\
\hline
Bank & 162 & 3159 & 3678 \\
\hline
Creditcard & 197 & 1278 & 1362 \\
\hline
Adult & 101 & 1371 & 1428 \\
\hline
Intrusion & 169 & 2962 & 3148 \\
\hline
\end{tabular}
\caption{Time comparison: Average training time (seconds) across various datasets.}
\label{tab:time}
\end{table}

According to Table \ref{tab:time}, FLIGAN shows a longer computational time compared to both FedAvg and FedGAN across most datasets, indicating a trade-off between time efficiency and accuracy. The increased time for FLIGAN is expected, given its comprehensive training process, including synthetic data generation. FedAvg, lacking GAN-based training, naturally exhibits the shortest computation times among the three, underscoring a trade-off between efficiency and the depth of model enhancement FLIGAN provides. FLIGAN may require more computational time but its ability to deliver superior model accuracy in certain scenarios highlights its potential as a preferred strategy,  especially in scenarios prioritizing prediction quality over computational resource constraints.

\subsubsection{\textbf{Analysis of  the effect of new synthetic data  in FL training}}

\begin{table}[!t]
  \centering
  
  \resizebox{\columnwidth}{!}{%
    \begin{tabular}{|l|c|c|c|c|}
      \hline
      \textbf{Dataset} & \textbf{Real data} & \textbf{Synthetic data} & \textbf{\% of new data} & \textbf{\# of steps} \\
      \hline
      Albert & 58k & 8k & 12\% & 5 \\
      Creditcard & 13k & 232 & 1\% & 1 \\
      Bank & 10k & 400 & 3\% & 2 \\
      Adult & 49k & 15k & 23\% & 8 \\
      Intrusion & 25k & 408 & 1\% & 1 \\
      \hline
    \end{tabular}
  }
  \caption{ Dataset statistics: It shows the amount of synthetic data added to achieve the best reported accuracy.}
  \label{fig:tab1}
\end{table}

\begin{figure}[h]
    \centering
    \includegraphics[scale=0.57]{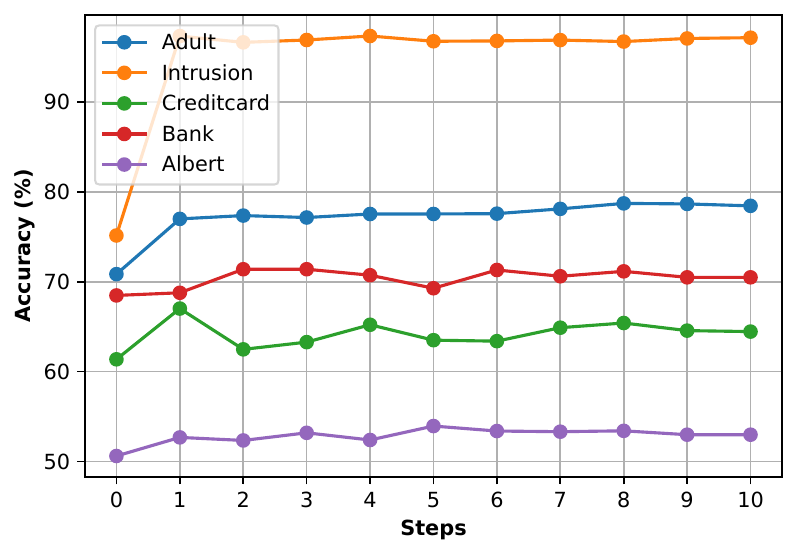}
    \caption{Step by addition of synthetic data:  The X-axis marks 1\% increments of GAN-generated data added to each node, based on the largest class sample size across nodes. The Y-axis shows the FL model's accuracy after adding data (step 0 as the baseline accuracy before any synthetic data integration).}
    \label{fig:steps}
\end{figure}

Figure \ref{fig:steps}, together with combined dataset statistics provided in Table \ref{fig:tab1}, reveal how the incorporation of synthetic data at a proportion affects the accuracy of a federated classification model across different datasets.
The \textit{Intrusion} dataset reaches maximum accuracy at Step 1 (75\% to 98\%), indicating the effectiveness of even a small amount of synthetic data. The \textit{Adult} dataset shows a notable initial improvement in accuracy with synthetic data, which slowly improves till Step 8 (70\% to 78\%). The \textit{Albert} dataset sees minimal gains from a larger synthetic data addition (4\% increase), suggesting a possible misalignment with the real data or bad quality of synthetic data. The \textit{Creditcard} dataset experiences an initial boost in accuracy (7\% increase), followed by a decline as more synthetic data is added, pointing to potential data quality inconsistency. The \textit{Bank} dataset's accuracy improves modestly and fluctuates after Step 2, implying additional synthetic data does not contribute as significantly to the model's learning. These findings underscore that even small amounts of synthetic data can enhance model accuracy, although the extent of improvement varies depending on the dataset.

\subsubsection{\textbf{ML efficacy analysis}}

\begin{figure}[!t]
    \centering
    \includegraphics[scale=0.53]{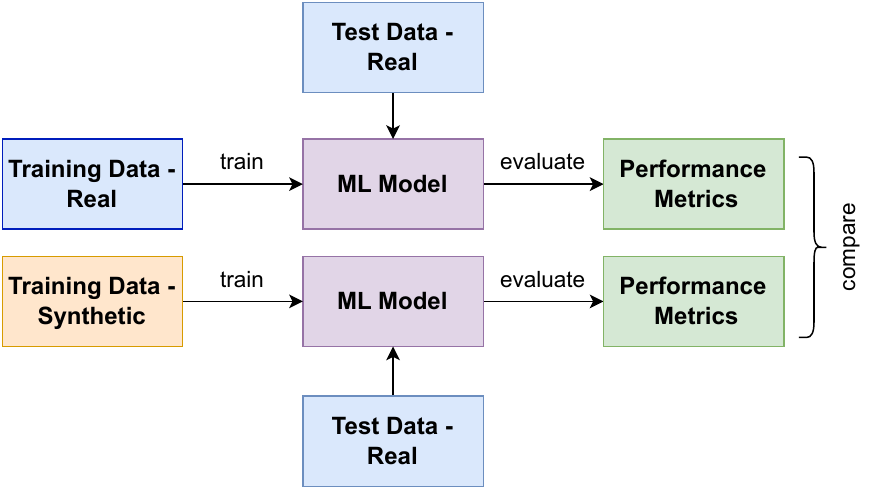}
    \caption{ML efficacy technique}
    \label{fig:MLE}
\end{figure}

We validated GAN-generated datasets against real datasets using a machine learning efficacy method \cite{xu2019modeling}, as shown in Figure \ref{fig:MLE}. We divided real and synthetic datasets into training and test subsets, and trained models with a Random Forest Classifier under identical settings for accurate comparison. Performance was evaluated on the real dataset's test subset to assess the training effectiveness of synthetic versus real data.

\begin{figure}[h]
    \centering
    \includegraphics[scale=0.57]{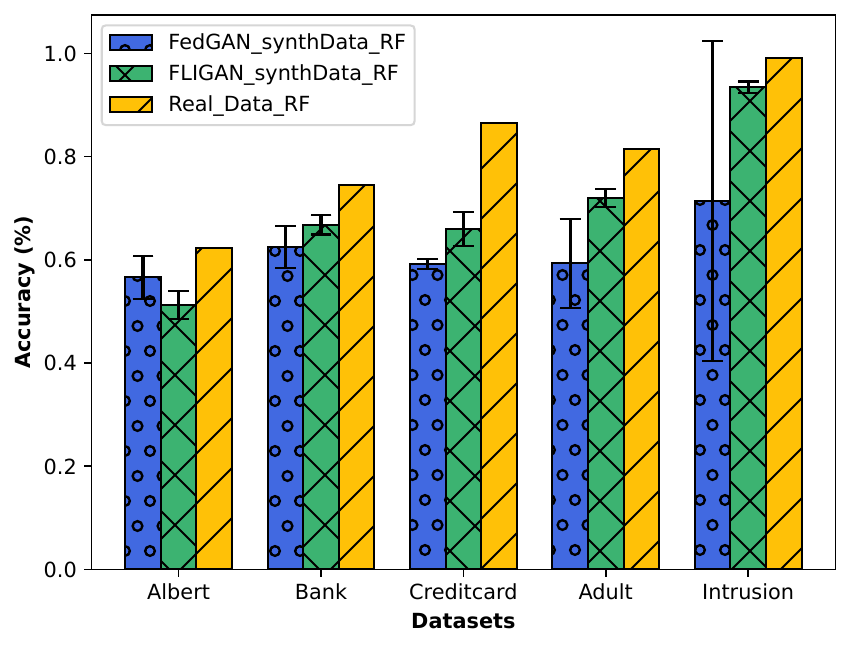}
    \caption{ML efficacy analysis: Compares quality of FedGAN and FLIGAN generated data against real data via Random Forest classifier, highlighting average accuracy across datasets. }
    \label{fig:eff}
\end{figure}

Figure \ref{fig:eff} shows FLIGAN-generated (\textit{FLIGAN\_synthData\_RF}) data nearly matches real data (\textit{Real\_Data\_RF}) in accuracy for Intrusion dataset, outperforming FedGAN (\textit{FedGAN\_synthData\_RF}) in effectiveness and consistency. This indicates FLIGAN's superior ability to produce high-quality synthetic datasets for FL, offering more representative and robust data with more consistent results.

\textbf{Discussion.} Our FL framework requires edge nodes to share metadata such as categorical columns' unique values and class label's metadata. We assume a secure mechanism exists for this, but metadata transmission during GAN training remains vulnerable to attacks. While common in FL frameworks, privacy concerns could be mitigated by future integration of encryption and privacy-preserving methods like homomorphic encryption \cite{dwork2006calibrating}, though this is beyond our current scope.

\section{Related Work}\label{sec:relatedwork}

FL algorithms like FedProx~\cite{li2020federated}, FedNova~\cite{wang2020tackling}, and SCAFFOLD~\cite{karimireddy2021scaffold} address the issue of non-IID data, each enhancing the model performance in unique methods. However, they fall short in extreme cases of incomplete data. Standard synthetic data generation techniques, such as Random Oversampling, SMOTE~\cite{chawla2002smote}, Cluster-Based Oversampling~\cite{jo2004class}, and Gaussian Mixture Model (GMM) ~\cite{chokwitthaya2020applying}, can address data incompleteness. However, these methods often struggle with complex data patterns.  Advanced deep learning approaches, especially GANs with architectures like CTGAN~\cite{xu2019modeling}, and TabFairGAN~\cite{rajabi2022tabfairgan}, offer solutions by effectively handling tabular data. But, in an FL environment, where data remains decentralized and privacy-preserving is necessary, these methods face limitations. 
Fed-TGAN \cite{9162672} and HT-Fed-GAN ~\cite{htfedgan} tackle non-IIDness of tabular data in FL with advanced privacy-preserving features, but face challenges like computational complexity of GMMs and complex parameter tuning. Table \ref{tab:algorithm_comparison} compares our work with other relevant studies. FLIGAN offers a simplified alternative, focusing on efficient synthetic data generation through classwise sampling and node grouping, reducing computational demands and simplifying the process while preserving data privacy in incomplete data scenarios.

\begin{table}[!t]
\centering
  \scriptsize
\caption{Related work: Comparison of most relevant works, that use synthetic data for enhancing FL}
\label{tab:algorithm_comparison}
\begin{adjustbox}{width=1\columnwidth}
\begin{tabular}{lcccc}
\hline
\multicolumn{1}{c}{\textbf{Algorithm}}                  & \multicolumn{1}{c}{\textbf{\begin{tabular}[c]{@{}c@{}}Privacy\\ preservation\end{tabular}}} & \textbf{\begin{tabular}[c]{@{}c@{}}Synthetic data\\ generation\end{tabular}} & \textbf{\begin{tabular}[c]{@{}c@{}}Deals with \\ incomplete data \end{tabular}} & \textbf{Method used}                                                                            \\ \hline
CTGAN \cite{xu2019modeling}            & X                                                                               & $\checkmark$                                                                         & X                                                                                        & GAN with VGM                                                                                    \\
TabFairGAN \cite{rajabi2022tabfairgan} & X                                                                               & $\checkmark$                                                                         & X                                                                                        & GAN                                                                                             \\
Fed-TGAN \cite{9162672}                & $\checkmark$                                                                    & $\checkmark$                                                                         & X                                                                                        & \begin{tabular}[c]{@{}c@{}}Federated GAN\\  with VGM\end{tabular}                               \\
HT-Fed-GAN \cite{htfedgan}             & $\checkmark$                                                                    & $\checkmark$                                                                         & X                                                                                        & \begin{tabular}[c]{@{}c@{}}Federated GAN\\  with VB-GMM\end{tabular}                            \\
\textbf{\begin{tabular}[c]{@{}l@{}}FLIGAN \\ (Our Work)\end{tabular}} & $\checkmark$                                                                                & $\checkmark$                                                                 & $\checkmark$                                                                     & \begin{tabular}[c]{@{}c@{}}Federated WGAN \\ w/ classwise sampling\\  \& node grouping\end{tabular} \\ \hline

\end{tabular}
\end{adjustbox}
\end{table}
\section{Conclusions and Future Work}\label{sec:conclusion}

In this study, we introduce FLIGAN, an efficient method using federated GANs to synthetically augment incomplete data for improved accuracy. We benchmarked FLIGAN against FedAvg and FedGAN across various datasets and showed improved accuracy, particularly with complex datasets like \textit{Intrusion} and \textit{Adult}. Although FedAvg is faster, FLIGAN's accuracy gains demonstrate its effectiveness in FL environments with incomplete data. In the future, we plan to refine FLIGAN for greater accuracy and broader compatibility, exploring advanced GAN architectures for different data types.

\section*{Acknowledgements}

This research was funded in part by the Austrian Science Fund (FWF) through following projects: \textit{Transprecise Edge Computing (Triton)}, Austrian Science Fund (FWF):  10.55776/P36870, \textit{Trustworthy and Sustainable Code Offloading (Themis)}, Austrian Science Fund (FWF): 10.55776/PAT1668223, and by the Austrian Research Promotion Agency (FFG) through the following project: \textit{Satellite-based Monitoring of Livestock in the Alpine Region (Virtual Shepherd)}, FFG Austrian Space Applications Programme ASAP 2022 \#53079251.

\bibliographystyle{unsrt}

\bibliography{references}
\end{document}